\documentclass[twocolumn]{article}
\usepackage[utf8]{inputenc}
\usepackage{amsmath}
\usepackage{authblk}
\usepackage{graphicx}
\usepackage{pgfplots}
\usepackage{pgf-pie}
\usepgfplotslibrary{
  groupplots
}
\graphicspath{ {./images/} }

\title{Evaluating the performance of the LIME and Grad-CAM explanation methods on a LEGO multi-label image classification task}
\author[1]{Cian, David}
\author[2]{van Gemert, Jan (supervisor)}
\author[3]{Lengyel, Attila (supervisor)}
\affil[1,2,3]{Delft University of Technology}
\date{\today}

\begin{document}

\maketitle

\begin{abstract}
    In this paper, we run two methods of explanation, namely LIME and Grad-CAM, on a convolutional neural network trained to label images with the LEGO bricks that are visible in them. We evaluate them on two criteria, the improvement of the network's core performance and the trust they are able to generate for users of the system. We find that in general, Grad-CAM seems to outperform LIME on this specific task: it yields more detailed insight from the point of view of core performance and 80\% of respondents asked to choose between them when it comes to the trust they inspire in the model choose Grad-CAM. However, we also posit that it is more useful to employ these two methods together, as the insights they yield are complementary.
\end{abstract}

\section{Introduction}

A dump of the raw data from an experiment isn't what you'd call a scientific article, and it certainly wouldn't be published in any reputable publication. Indeed, scientific papers never directly include the raw data, and for good reason: first, it would run for hundreds of pages, but more importantly, it would be virtually useless to the reader. Of course, the raw data should be available, for reproducibility purposes, but the article itself is instead a structured and meaningful presentation of results stemming from the data. The claims made in a paper must be both clear and supported by the data, and the universally adopted compromises to achieve this are tables and figures.

This is a prime example of the main ideas behind explanations. The brain's cognitive capabilities are limited, and it has an easier time understanding data in certain formats than others. In turn, understanding a phenomenon, which is to say having a mental model thereof, is a platform for further deductive reasoning about the phenomenon, which makes understanding a useful asset for a variety of tasks. In general, we refer to any technique that affords us greater understanding as an explanation technique. No wonder then that, as a technology which garners a lot of attention, while being rather complex, we desire artificial intelligence (AI) to be explained to us. Indeed, this has given birth to a new subfield attempting to explain not the theoretical underpinnings of AI but the behavior of precise instances of AI systems: explainable AI (XAI).

Deep learning (DL) is a paradigm of AI which makes use of deep artificial neural networks (ANN). An ANN is collection of computing units inspired by biological neurons and organized in layers, where one layer feeds input to the next. The first layer is an input layer, while the last layer is an output layer. Strictly speaking, when there are one or more hidden layers between the input and output layer, we have a deep learning network. These models, which are often referred to as machine learning (ML) instead of AI, come with many trainable parameters. In fact, the amount can range in the hundreds of millions. As such, even with access to the parameters, no human can understand the operation of the network without further processing and aggregation of these parameters into a representation that is easier to digest.

Not all explanations are created equal, and it is important when choosing an explanation to pick the right tool for the job. Different stakeholders have different expectations as to a deep learning model. Some, such as candidates being subjected to a DL-powered hiring system, might expect it to not show any racial or gender bias. Others, such as the military, might expect their systems to be safe from malicious actors. Depending on the explanation technique chosen, we can expect different performance depending on the metrics we look at. Looking at the terms in a candidate's CV that the DL model considers most relevant to the hiring decision is an easy way of checking for race or gender bias. On the other hand, this isn't necessarily what the military is looking for: instead, they might find data about the model's sensitivity to minor perturbations in the input data, as are common in evasion attacks on DL models, to be better.

The question we aim to answer in our paper is: how does the performance of explanation techniques for deep learning compare on a multi-label classification task? Multi-label classification consists of assigning multiple labels to an input. In our case, the precise task is one that is painstakingly performed on a daily basis by children all around the world, yet oddly overlooked in research literature: namely, when looking at a pile of LEGO bricks, figuring out if it contains that one piece you're missing for your precious LEGO Star Wars AT-AT. Concretely, we want to label an image of a pile of LEGO bricks with the names of all the visible bricks.

There are two main benefits to applying explanation techniques to LEGO brick identification in images. First, an enhanced understanding of the inner workings of the DL model allows the developers of the model to improve core performance, which is the umbrella term we use to denote all metrics relating to how well the model labels the images. Second, it can strengthen the trust that users place in the DL model's results, which is crucial, as people prefer using systems they trust. As such, we will evaluate the performance of the models on these two criteria.
We bring the following contributions. First, we apply two state-of-the-art explanation methods for deep learning networks to the LEGO identification task. Second, we discuss and compare their performance on several metrics pertaining to the two criteria of improving core performance and improving trust.

\section{Explanation theory and related work}

We drew inspiration for the LEGO identification context of the paper from a contraption built by Daniel West to automatically sort LEGO bricks by type into separate bins \cite{west_worlds_nodate}. Previous research comparing two explanation methods experimentally can be found in \cite{samek_explainable_2017}, which compares sensitivity analysis (SA) and layer-wise relevance propagation (LRP).

\subsection{Explanation scope}

Broadly speaking, explanation methods as they exist today belong to two categories in terms of scope: global and local explanations \cite{doshi-velez_towards_2017}. Global explanation methods provide information about the model itself, irrespective of any particular input: explanation methods in this category are akin to inductive reasoning. Local explanation methods, by contrast, examine a model's inference process based on a specific input in order to explain its prediction: they are akin to abductive reasoning. In \cite{gilpin_explaining_2018}, these methods are said to explain respectively the data \textit{representation} a network has learned and its \textit{processing} during an instance of inference.

\subsection{Faithfulness and interpretability}

The faithfulness of an explanation method refers to how accurately it portrays a model's functioning. Its interpretability refers to how easy it is for a human to improve his or her understanding of the model's function by using the explanation method. While not mutually exclusive, improving faithfulness usually increases complexity, whereas improving interpretability usually decreases complexity. Choosing between faithfulness and interpretability isn't a zero-sum game, as certain methods do better than other on both fronts, but this is a major trade-off. Indeed, \cite{selvaraju_grad-cam_2020} picks up on the faithfulness-interpretability trade-off and concludes that Grad-CAM offers both faithfulness and interpretability. This trade-off is also outlined in \cite{gilpin_explaining_2018}, where faithfulness is referred to as \textit{completeness} instead.

\subsection{Evaluation metrics}

As outlined in \cite{doshi-velez_towards_2017}, metrics for the evaluation of explanation methods can be divided into three categories: application-grounded, human-grounded and functionally-grounded metrics. Application-grounded metrics measure the success of humans on real-life tasks with intrinsic value, when aided by an explanation technique: for instance, such a metric could be obtained by comparing the performance obtained by a deep learning network tuned by an expert with and without access to the explanation technique. Human-grounded metrics rely on human judgment to evaluate methods on generally desirable criteria rather than criteria specific to a certain task. A simple such human-grounded metric, also mentioned in \cite{doshi-velez_towards_2017} is binary forced choice, where the humans participating to the evaluation of the methods are presented with the explanations produced by two methods and have to choose that which they consider of higher quality. Finally, functionally-grounded metrics evaluate explanation methods without any human input: a basic metric sometime used in this category is model sparsity when using lasso (L2) regularization, as described in \cite{lipton_mythos_2017}.

\section{Method}

\subsection{Kernels and feature maps}

Convolutional neural networks are called this way due to their use of convolutional layers which apply convolutions to their input. An image can be represented as a three dimensional tensor, where the last dimension, for instance, can represent the three channels, namely red, green and blue for an RGB image. The convolution is computed as the sum of the elements of a region of the image weighted by the coefficients of another three dimensional tensor called the convolution kernel.

In conventional computer vision methods, convolution kernels such as the Prewitt or Sobel filters have fixed weights and are designed by humans. In CNNs, the principle behind convolution stays the same, but the network learns the weights of the kernels through backpropagation.

Visualizing the learned convolution kernels makes for a rudimentary method of explanation. It is also a perfect illustration of the faithfulness for interpretability trade-off. On one hand, the method is a very faithful one, as it displays the network's convolutional layers without any loss of information: all the weights are shown. On the other hand, it is hard to drawn anything but basic conclusions from this explanation. For instance, it is fairly easy to see when a kernel for line (or edge) detection has been learned on the first convolutional layer, but kernels on the first convolutional layer can also learn kernels tailored to the specific input distribution, which aren't readily interpretable for humans. Visualizing kernels on deeper layers yields almost no interpretable information, as the convolutions on deeper layers are learned so as to work with the learned convolutions from the previous layer which feed into them.

Visualizing convolution kernels is the only global explanation method we use in this paper. Moving into the realm of local explanations using convolution kernels is a useful complement to the purely global method of looking strictly at the kernels themselves. Any visualization of the output of an entire hidden layer, such as a fully connected layer of perceptrons, is called an activation map. In the case of a convolutional layer, the activation map is also called a feature map, as convolutional networks can be considered to learn visual features of an image. Visualizing these feature maps for an input image provides a workaround for the problem encountered with kernel visualization, where learned kernels on a layer are adapted to the learned kernels of the previous layer, as it directly shows the response of a layer to its input. For stacked convolutional layers, the feature maps maintain the locality of activations, which means that the response of a deep convolutional layer with respect to the input image can be seen. Pooling layers lead to loss of locality due to the downsampling of their input, but for small window sizes, such as 2 by 2, this effect is negligible.

In this paper, we visualize both convolution kernels and feature maps in the early stages of our analysis, due to their simplicity and faithfulness to the underlying model. However, these methods constitute little more than a smoke test of the model under analysis.

\subsection{Gradient-weighted class activation mapping (Grad-CAM)}

Grad-CAM is an explanation technique which can be applied to any CNN post-hoc, without requiring any modification of the network under analysis \cite{selvaraju_grad-cam_2020}. It is a generalization of class activation mapping (CAM) \cite{zhou_learning_2015}, which by contrast requires a network architecture consisting of stacked convolutional layers, followed by global average pooling (GAP) and a softmax activation. For simplicity, we present it only in the context of a CNN for image classification. When given an input and a class, Grad-CAM produces a heat map of the class activation: it color codes parts of the input image according to how much they positively contribute to the system labeling the image with the provided class. In other words, it produces a class activation map.

First, the gradient $\frac{\partial y^c}{\partial A^k}$ of the score $y^c$ for class $c$ before the final soft-max layer is computed with respect to the feature map activation $A^k$ of convolution $k$ from the deepest convolutional layer. Then, the backpropagated gradients are globally max-pooled to yield the importance of the convolution for the class, $\alpha^c_k$: this is called the neuron importance weight, hence the adjective "gradient-weighted". Finally, a localization map $L^c$ is obtained by computing the weighted sum, followed by a ReLU, to obtain only positive contributions.

$$L^c = ReLU(\sum_k \alpha^c_k A^k)$$

\subsection{Local interpretable model-agnostic explanations (LIME)}

LIME is an explanation technique which is, as the name suggests, model-agnostic \cite{ribeiro_why_2016}: it can be applied mutatis mutandis to any kind of classifier. For a given input and prediction, it generates a readily interpretable model which is locally close to the real model. Our LEGO brick labeling is a multi-label image classification task, which in turn is just a series of binary image classification tasks. For such a task, LIME generates a linear model which takes as input superpixels of the original image. The superpixels are computed using the quickshift algorithm \cite{vedaldi_quick_2008}. Then, a heatmap can be overlayed over the input image, highlighting superpixels that contribute positively to a prediction in green and superpixels that contribute negatively in red. The highlighted superpixels can be selected if their contribution to the prediction is over a certain threshold or as the top $N$ superpixels that contribute the most (positively or negatively).

\section{Experiments}

\subsection{Data}

The data used to train the network is made up of 819 pictures of resolution 4608 by 3072 and 817 pictures of resolution 5472 by 3648. In these pictures, between 1 and 10 LEGO bricks are visible, drawn from a LEGO box containing 85 different types of bricks. Lighting conditions vary: pictures can be illuminated by natural or artificial light. Backgrounds also vary, ranging from plain backgrounds consisting of only one solid color to intricate textures. In addition, synthetic data was used in preliminary stages to explore viable network architectures. The synthetic data was generated by running a headless Blender on a GPU compute cluster.

\subsection{Convolutional neural network}

To label images with the visible LEGO bricks, we use a convolutional neural network (CNN), inspired by the architecture of AlexNet \cite{krizhevsky_imagenet_2017}.

\begin{figure}[h]
\centering
\includegraphics[width=8cm, height=1.86cm]{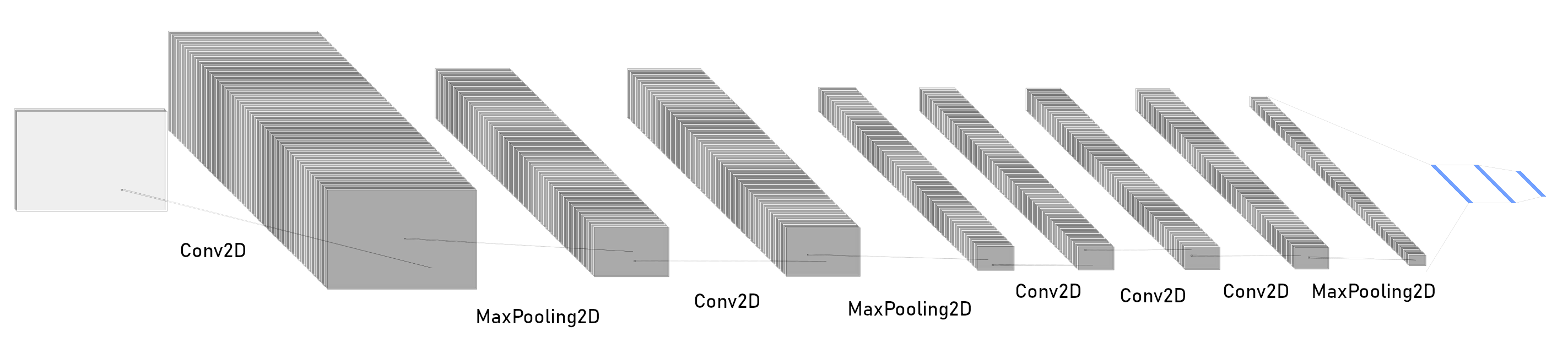}
\caption{Network architecture. The network consists of five convolutional layers and three dense layers, with three maximum pooling layers and three batch normalization layers in the convolutional part of the network.}
\end{figure}

We use leaky rectified linear units (leaky ReLU) as an activation function for all hidden layers. Leaky ReLU is a variant of the standard rectified linear units (ReLU). The ReLU activation function is defined as such: $f(x) = \text{max}(0, x)$. It is a non-saturating non-linearity. In this context, non-saturating means that as $x$ goes toward infinity, the activation function also goes toward infinity (as opposed to say, the hyperbolic tangent activation function which squeezes the input between $-1$ and $1$). ReLU is prone to the "dying problem", where negative input will yield a null gradient from which it is hard to recover, leading to large parts of the network becoming useless over time. Leaky ReLU, on the other hand, has a slope of $\alpha$ for negative input, which mitigates the dying ReLU problem. This is important for our comparison of explanation methods because many of them, such as Grad-CAM, make use of gradients: large swathes of dying neurons can in some cases cause numerical instability, and otherwise yield technically correct, but unremarkable explanations consisting of almost uniformly zero intensity heatmaps.

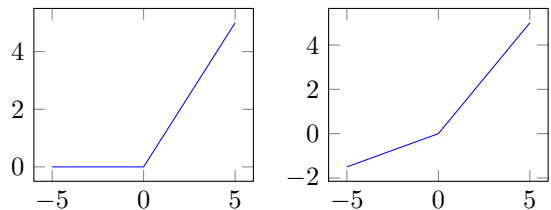
\begin{figure}[h]
\centering
\begin{tikzpicture}
  \begin{groupplot}[group style={group size=2 by 1}]
    \nextgroupplot[width=4.5cm, no marks]
      \addplot {max(0, x)};
    \nextgroupplot[width=4.5cm, no marks]
      \addplot {max(0.3 * x, x)}; 
  \end{groupplot}
\end{tikzpicture}
\caption{ReLU and leaky ReLU. Leaky ReLU maintains a small slope for negative inputs, mitigating the problem of dying neurons.}
\end{figure}

During training, we use an Adam optimizer and a custom loss function: weighted binary cross-entropy with logits. The training dataset consists of images where between $1$ and $10$ LEGO bricks are visible, which can belong to $85$ different classes of bricks. This means that in any given image, roughly between $1.2\%$ and $12\%$ of the labels appear in the image annotation. As such, when using a regular binary cross-entropy function, the network easily learns to output no labels, since it is much more prone to outputting false positives than false negatives. In order to circumvent this, our weighted binary cross-entropy with logits is defined as follows:

$$\text{W-BCE}(x) = -t \log(x) w_{\text{positive}} - (1 - t) \log(1 - x)$$

where $x$ is the prediction in logits for a single label, $t$ is the target label ($1$ if brick is visible, $0$ otherwise), and $w_{\text{positive}}$ is the coefficient by which we weight positive predictions to increase recall and decrease precision (or vice-versa).

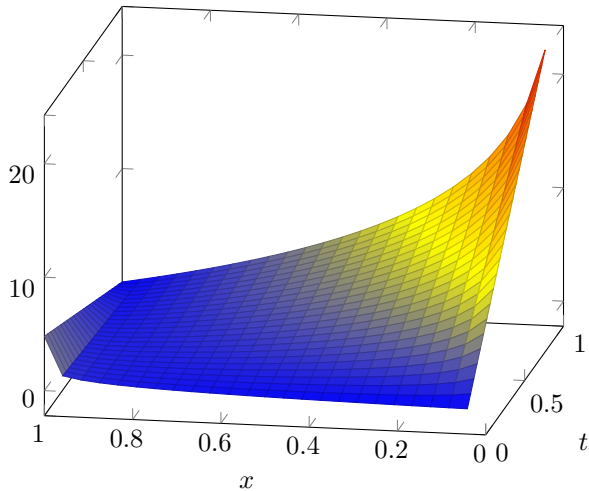
\begin{figure}[h]
\centering
\begin{tikzpicture}
  \begin{axis}[no markers, view={-80}{20}, xlabel={$t$}, ylabel={$x$}]
    \addplot3 [domain=0:1, surf] {-x * log10(y) * 16 - (1 - x) * log10(1 - y)}; 
  \end{axis}
\end{tikzpicture}
\caption{Weighted binary cross-entropy with logits. The graph is for a positive weight of 16. Looking at the graph for $(t, x) = (1, 0)$, which represents a false negative, and $(t, x) = (0, 1)$, which represents a false positive, the loss is much greater for false positives than false negatives.}
\end{figure}

\subsection{Core performance}

The specific problem we focus on is to label an image with the LEGO bricks visible in it. As a multi-label classification task, the problem belongs to a common class of tasks. However, to the knowledge of the author, no research except that concerning the construction of the LEGO sorting machine has been done into this specific instance. Furthermore, several factors contribute to making this a fairly hard problem in the realm of multi-label classification. During the development and training of our network, we will analyze it using the chosen explanation methods in order to see what information we can glean about the model's functioning and in some cases mitigate these factors.

It is quite common, especially for novices lacking extensive experience with machine learning models, to miss certain less obvious aspects of a problem which nonetheless play a major role. In the case of LEGO multi-label classification, a plausible error when overlooking the class imbalance caused by a brick being much more often absent than present in an image is to use a plain binary cross-entropy (BCE) loss function. This causes the network to quickly converge toward outputting all bricks as being always absent from the images. Taking a quick glance at the output of Grad-CAM, we see that none or almost none of the image contributes positively to the predictions when training the network with BCE. We note however that this could have been deduced just as well by looking at the learned kernels or the LIME explanation with a very low threshold.

\begin{figure}[h]
\centering
\includegraphics[width=2.4cm, height=1.2cm]{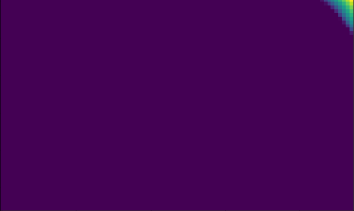}
\caption{Grad-CAM output for network trained with BCE. Except the top right corner, no part of the image contributes positively to the prediction for a certain brick being present with a confidence of almost zero (in other words, the brick is predicted as absent with high confidence).}
\end{figure}

We examine now the network after 40 epochs of training with a weighted binary cross entropy with logits loss function. It registers a loss of 16.251 (with false negatives set to count 22 times as much as false positives), and a binary accuracy of 84,71\%. Looking at the convolution filters learned by the first convolutional layer, we can make several remarks. For the first layer only, since the filters are three dimensional tensors, we can visualize them in two dimensions as RGB images.

\begin{figure}[h]
\centering
\includegraphics[width=7cm, height=3cm]{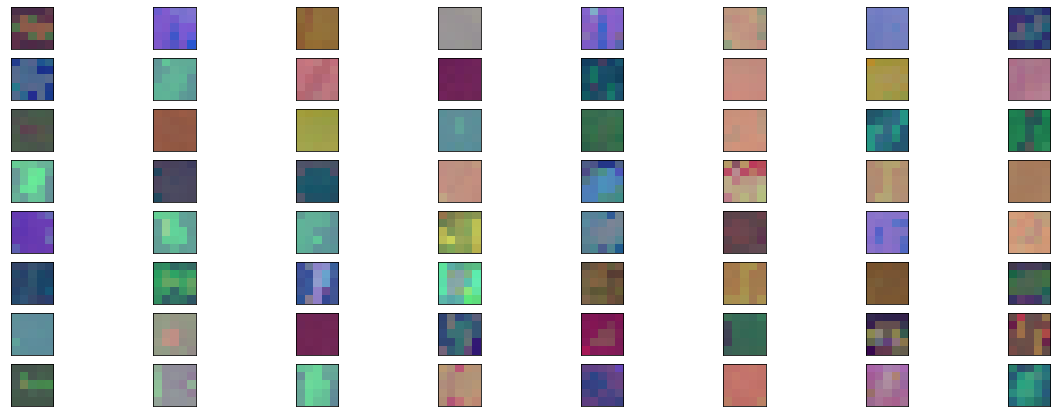}
\caption{Learned convolution kernels after 40 epochs. Only one fourth of all the learned kernels are shown here.}
\end{figure}

As we might expect, the network learns several kernels which seem to be thin line detectors or partly learned edge detectors. These kinds of kernels are often encountered when engineering features by hand in computer vision, so it is reassuring to see the network make use of these kernels.

\begin{figure}[h]
\centering
\includegraphics[width=4.25cm, height=1cm]{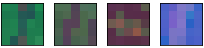}
\caption{Partly learned edge detectors. The edge detection kernels are not as clean as those defined by a human, and we cannot thus be sure that they will evolve into effective edge detectors.}
\end{figure}

We also see on the kernel visualizations that many of them consist of a more or less uniform color, and that such filters cover a wide range of colors. This is also reassuring, as one might expect the network to discriminate between bricks heavily based on their color: indeed, while it doesn't allow the network to decide between two bricks of the same color but different shape, it is an easy criterion to learn to at least distinguish between bricks of different colors.

For deeper convolutional layers, there is no easily interpretable visualization. We show a glimpse of the second convolutional layer's kernels to allow the reader to convince himself or herself of the poor interpretability of kernel visualizations for layers beyond the first.

\begin{figure}[h]
\centering
\includegraphics[width=4.25cm, height=3.3cm]{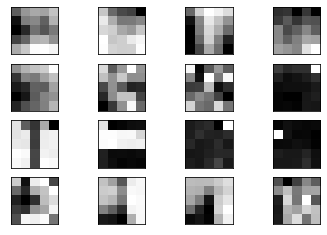}
\caption{Learned kernels of the second convolutional layer. On each line are displayed the first four channels of a kernel.}
\end{figure}

Moving on to feature maps, as computed for a specific input image, we can make a couple of observations. As we go deeper along the convolutional stack, we see that the network focuses more on the bricks and less on the background: in the last convolutional layer, most feature maps activate almost exclusively on bricks. However, it is the other way around for a minority of convolutions, as they activate for the background instead. Another interesting observation is that the network distinguishes between bricks, as can be seen from certain feature maps where one or more bricks produce large activations as opposed to other bricks, but it globally still has a hard time distinguishing between them, as in most activation maps all bricks are highlighted with similar intensity. A potential cause for this result is the similarity of certain bricks, which causes a relatively untrained network such as ours, which was only trained for 40 epochs, to not be able to discriminate between bricks.

\begin{figure}[h]
\centering
\includegraphics[width=4.7cm, height=8.5cm]{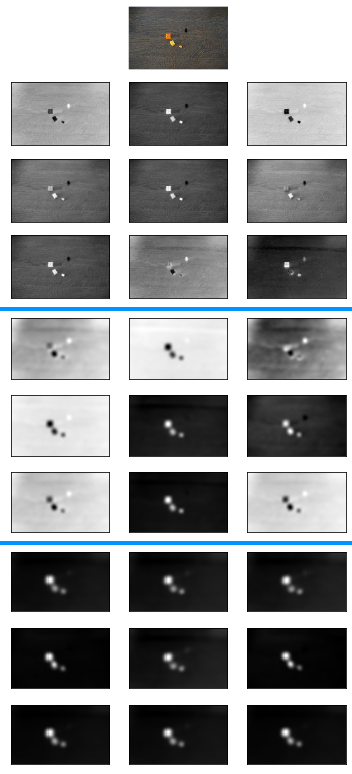}
\caption{Activation maps. The image at the top is the input image. Below are shown 9 activation maps from respectively the first, third and fifth (and last) convolutional layers.}
\end{figure}

Visualizing kernels and feature maps can reveal interesting information, but it makes for lackluster explanation. A major shortcoming is that these methods do not tell us anything about the functioning of the dense part of the network, or any other layers for that fact except the convolutional layers. In short, they only tell a small, isolated part of the story behind a prediction. For this reason, we turn to more advanced, holistic methods: Grad-CAM and LIME.

\begin{figure}[h]
\centering
\includegraphics[width=8.2cm, height=1.7cm]{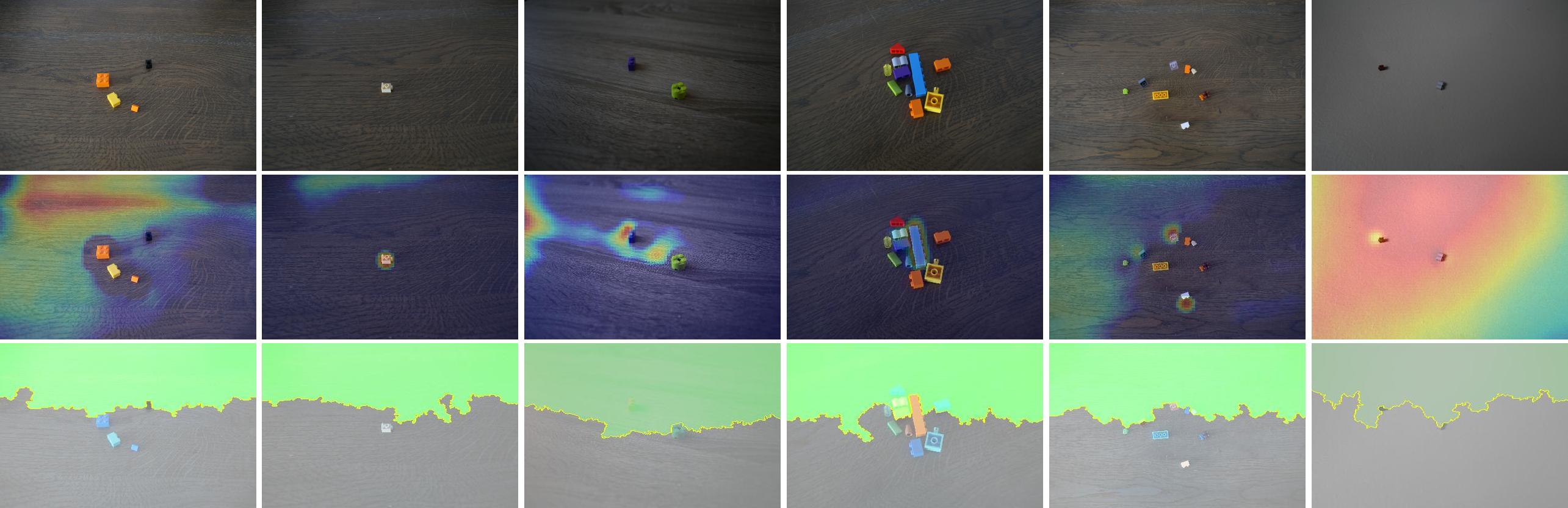}
\caption{Grad-CAM and LIME explanations. The top row shows six input images. The middle row shows the Grad-CAM generated explanations, for the brick predicted with the most confidence as being present. The bottom row shows the LIME generated explanations, displaying the 40 superpixels which contribute the most (positively) to a prediction.}
\end{figure}

The Grad-CAM explanations confirm our previous remark that the network sometimes looks at the background instead of the bricks, sometimes has trouble distinguishing between bricks and looks at all of them together, and sometimes looks precisely at one brick. This is where the class-discriminativity of Grad-CAM comes in handy, as it clearly shows when the network can differentiate between the bricks it sees. On the other hand, LIME doesn't seem to generate particularly telling explanations in this case. LIME does in general offer however two benefits compared to Grad-CAM: it builds a linear model, from which easy to understand and concisely expressed knowledge is much easier to extract, and it highlights superpixels, which means it is possible to see how the network reasons based on patches of similar pixels.

\subsection{Trust}

We use binary forced choice to compare the trust in the network inspired by Grad-CAM and LIME. We do not consider kernel visualizations and feature maps for trust, since understanding these requires moderate expertise in deep learning for computer vision. 10 participants must answer 10 binary forced choice questions, where they are provided for each with the original input image the two explanations generated by Grad-CAM and LIME and must choose between the two according to which method causes them to trust the system more. The test is conducted as a blind study, where the participants are not informed of which method is which. Additionally, they are not informed of the model's prediction in order to avoid the correctness of the prediction swaying their vote. The participants must also qualify their level of familiarity with deep learning, in order to check if familiarity changes how trustworthy a model is perceived based on the explanations. 

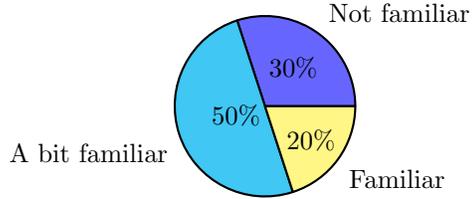
\begin{figure}
\centering
\begin{tikzpicture}
    \pie[radius=1.2]{30/Not familiar, 50/A bit familiar, 20/Familiar}
\end{tikzpicture}
\caption{Level of familiarity with deep learning of survey respondents.}
\end{figure}

On average, $80\%$ of respondents find Grad-CAM to inspire more trust, and $20\%$ find LIME to inspire more trust. Based on our discussion of the core performance criterion, this result confirms our remark that Grad-CAM seems more adapted for this task, network architecture, and stage of training across users of varied levels of expertise.

\subsection{Reproducibility}

We strive to make our experiments easy to check and reproduce. To this end, we invite the reader to view configuration parameters (e.g. random number generator seeds for NumPy and TensorFlow), hyperparameters, model architecture, weights and performance across runs on Weights \& Biases \cite{cian_weights_nodate}. The Jupyter notebook used for the project, as well as several utility scripts are publicly available on GitLab \cite{cian_gitlab_nodate}. The part of the real dataset which was used to train the networks, consisting of about 1800 images, is also available on a distant server from TU Delft, on request. Additionally, for the readers interested in generating more synthetic data or generating custom synthetic data, the code is also available on GitLab, on request. 

\section{Discussion}

In this paper, we evaluated two explanation methods, Grad-CAM and LIME, as well as kernel and feature map visualizations, on a convolutional neural network designed to label images with the LEGO bricks they contain. We conducted this evaluation on two criteria, namely the improvement of the core performance of the model, as well as the trust explanation methods can generate for the network under analysis. We saw that each method has its strengths and weaknesses, and that certain methods are more adapted than other for this use case, and for certain purposes: for instance, visualizing kernels is perfectly fine for gleaning basic insight, but a more complex analysis warrants the use of Grad-CAM or LIME. The conclusion we draw from our analysis is two-fold. First, we reiterate that one must choose the right tool for the job: not all explanation methods work equally well, and their performance varies depending on three factors: the objectives of the ML model (not just the core performance but also trust in this case), the users of the explanation (scores for trust Grad-CAM and LIME vary depending on the level of expertise), and the ML model under analysis. Second, it is wise to use multiple methods, as it is unlikely that only one method will yield all the insight that can be acquired.

Of course, this research was rather limited in scope due to time constraints. Only two out of the many existing methods of explanation were tried. Furthermore, methods of explanation belong to a wide and varied range of categories, and although Grad-CAM and LIME belong to two different categories, namely salience methods and linear proxy methods, other methods, such as activation maximization, would also have been interesting to include. The evaluation was conducted only qualitatively and was not formal, a shortcoming which is as is lamented enough in the research literature in the field of XAI. 

\section{Acknowledgments}

The author would like to acknowledge the generous support of this project's two supervisors, Prof. Jan van Gemert and Attila Lengyel, as well as the important part Berend Kam, Hiba Abderrazik, Nishad Thakur, and Rembrandt Oltmans played in the data collection and labeling. Finally, the author expresses his gratitude to the Delft University of Technology for the opportunity to pursue this research.

\bibliographystyle{unsrt}
\bibliography{XAI-research-project.bib}

\end{document}